\documentclass[conference]{IEEEtran}

\usepackage{amsfonts}

\usepackage{booktabs}
\usepackage{dcolumn}
\usepackage{siunitx}

\usepackage[
singlelinecheck=false 
]{caption}

\usepackage{cite}
\usepackage{hyperref}
\hypersetup{
    colorlinks=true,
    linkcolor=black,
    filecolor=magenta,      
    urlcolor=cyan,
    citecolor=blue,
}

\ifCLASSINFOpdf
   \usepackage[pdftex]{graphicx}
  \DeclareGraphicsExtensions{.pdf,.jpeg,.png}
\else
\fi

\usepackage{amsmath}
\usepackage{mathtools}

\usepackage{mathdots}
\def\vdots{\vbox{\baselineskip=5pt \lineskiplimit=0pt 
\kern6pt \hbox{.}\hbox{.}\hbox{.}}}

\usepackage{amsmath,nicematrix}

\usepackage{array}
 \usepackage[caption=false,font=footnotesize]{subfig}

\usepackage{float}
\usepackage{threeparttable}
\usepackage{multicol}
\usepackage{multirow}
\begin{document} 

\title{LiDAR data acquisition and processing\\ for ecology applications}

\author{\IEEEauthorblockN{Ion Ciobotari }
\IEEEauthorblockA{\small INESC-ID Lisboa\\
Instituto Superior Técnico \\
Universidade de Lisboa, Portugal\\
{\small 	ionciobotari@tecnico.ulisboa.pt }
}
\and
\IEEEauthorblockN{Adriana Príncipe and Maria Alexandra Oliveira}
\IEEEauthorblockA{\small Centre for Ecology, Evolution and Environmental Changes \\
CHANGE - Institute for Global Change and Sustainability, \\
Faculdade de Ciências\\
Universidade de Lisboa, Portugal \\
{\small \{aprsilva, maoliveira\}@ciencias.ulisboa.pt}
}

\and
\IEEEauthorblockN{João Nuno Silva}
\IEEEauthorblockA{\small INESC-ID Lisboa\\
Instituto Superior Técnico \\
Universidade de Lisboa, Portugal\\
{\small joao.n.silva@inesc-id.pt}
}
}

\maketitle

\thispagestyle{plain}
\pagestyle{plain}

\begin{abstract}
The collection of ecological data in the field is essential to diagnose, monitor and manage ecosystems in a sustainable way. Since acquisition of this information through traditional methods are generally time-consuming, due to the capability of recording large volumes of data in short time periods, automation of data acquisition sees a growing trend. Terrestrial laser scanners (TLS), particularly LiDAR sensors, have been used in ecology, allowing to reconstruct the 3D structure of vegetation, and thus, infer ecosystem characteristics based on the spatial variation of the density of points. However, the low amount of information obtained per beam, lack of data analysis tools and the high cost of the equipment limit their use. This way, a low-cost TLS (<10k€) was developed along with data acquisition and processing mechanisms applicable in two case studies: an urban garden and a target area for ecological restoration. The orientation of LiDAR was modified to make observations in the vertical plane and a motor was integrated for its rotation, enabling the acquisition of 360° data with high resolution. Motion and location sensors were also integrated for automatic error correction and georeferencing. From the data generated, histograms of point density variation along the vegetation height were created, where shrub stratum was easily distinguishable from tree stratum, and maximum tree height and shrub cover were calculated. These results agreed with the field data, whereby the developed TLS has proved to be effective in calculating metrics of structural complexity of vegetation.
\end{abstract}


%
\IEEEpeerreviewmaketitle

\section{Introduction}

Ecology is the science that studies the interaction between organisms and the biophysical environment they inhabit, including both biotic and abiotic components. Ecology studies require sampling field data, which demand intensive labour and time consumption for its collection. Therefore, methods and/or technologies, such as remote sensing, are bringing new opportunities to collect higher amounts of data with varying resolutions, but generally over areas which are paramount for ecology studies.

Light Detection and Ranging (LiDAR), based on laser light emission, belonging to the visible and near infrared electromagnetic spectrum, calculates the distance to an observed area or target by measuring the delay between emission and detection of each pulse. This technology can be separated in two classes according to its range measurement method: full waveform and discrete returns \cite{sumnall2016comparison} . Discrete return LiDAR returns a static number of amplitude measurement. Each return is registered when the reflected signal exceeds a threshold. Full waveform LiDAR returns an entire time-varying amplitude di, since it's easier to implement and returned data is easier to interpret.

When applied in ecology, LiDAR shows great potential in acquiring large amount of information of vegetation structure which has been applied in various case studies, such as: collecting information to evaluate forest standard dendrometric parameters (i.e. stem diameters, tree height, stem density, basal area and commercial wood volumes), canopy characterisation (i.e. canopy cover and gap fraction, leaf area ratios and foliage distribution) and tree structure modelling \cite{lim2003lidar, dassot2011use, disney2019terrestrial}, monitoring multiple ecological restoration objective outcomes \cite{almeida2019effectiveness}, determination of carbon biomass production \cite{edson2011airborne}, and quantification of vegetation structure complexity using height diversity indices \cite{listopad2018effect, listopad2015structural, fisher2014savanna}. Vegetation structure is considered as a primary determinant of habitat quality, defining the distribution and abundance of critical habitat components for animals, such as food, nesting sites, shelter, cover against predation and camouflage \cite{aguilarmarkus}.

There are many types of LiDAR systems, which can be generalised in three major platforms: spacial, aerial and ground systems. Beland et al. \cite{beland2019promoting} characterised some of these systems in terms of area coverage, resolution and main area of occlusion. Area coverage refers to the typical spacial area that the system is able to capture. Resolution refers to the level of detail resolved from LiDAR measurement, typically presented as the minimum distance between two points or as the minimum angle between laser firings. The lower these values are, the higher the resolution and detail of the point cloud.  Occlusion refers to blocking or shadowing of laser pulses by the target area of object, reducing the amount of information that can be extracted. By using these parameters to characterise LiDAR systems it's possible to discern their advantages and limitations. Doing so allows an user to choose the system best suited for his needs.

Aerial LiDAR systems (ALS), operating at high altitudes, are capable of high area coverage, however they also possess large footprints and low resolution. Newer aerial models are implemented on drones, allowing data capture at lower heights reducing the footprint and increasing image resolution. Ground or terrestrial systems (TLS), stationary or mobile, have the lowest and greatest values in area coverage and resolution, respectively. Mobile terrestrial systems (MLS), when compared with their stationary counterpart, are capable of greater area coverage. However, due to mobility, less time is spent on each visible target, reducing the system resolution, and are more prone for measurement errors. Occlusion of all of these systems depends not only on the system general location (ground, above ground, below and above canopy levels), but also on vegetation density of the study area.

Due to it's potential, LiDAR technology sees great improvements, not only on its increase of measurement accuracy and range and quantity and quality of information, but also on its various platform implementations. However, with this continuous growth, the LiDAR systems show more complex implementations and cost. For example, stationary LiDAR system have a market price starting at around 20000\$.

This way, the main purpose of this work is to develop an easy to build and reliable TLS. This system must be low cost, capable of capturing point clouds with height precision and resolution. This thesis presents the architecture and implementation of such system, which is mainly composed by changing the LiDAR sensor visual orientation and, with the help of supporting components, to both increase its resolution and to achieve the desired field of view. This solution's total cost did not exceed a value of 10000€, lower than the cheapest TLS presented previously. The developed system was also tested in various environments, where it showed great point cloud detail and measurements with resolution equivalent to LiDAR systems with a price of 20000€. Since the developed system showed promising results, there may be potential for the implementation of a similar system using much cheaper LiDAR sensors, further lowering the system cost and increasing application of this technology.

\section{Low cost LiDAR system (LCLS) arquitecture}




Excluding the LiDAR sensor, the system can be divided in four parts, the hardware components and actuators responsible to carry and change the position of the LiDAR sensor, the actuator control, sensor acquisition and processing. A turn table is a simple hardware solution, capable to hold the LiDAR sensor and rotates it in the horizontal plane with the use of a step motor. The step size and number of steps taken per rotation of the step motor can be configured in order to increase the data acquisition and resolution in the horizontal plane. Besides the LiDAR sensor, this table can also be used to mount additional sensors deemed necessary to increase image quality or correction. 




The LiDAR sensor provides the most part of sensor reading, more specifically, distance measurements to targets. The rest of sensor block is composed of location and movement tracking sensors. The actuators are responsible for moving the LiDAR sensor according to the instructions provided by the processing unit. This unit generates new instructions according to assessments of the data provided by the sensor block.


\begin{figure} [htb]
    \centering
    \includegraphics[width=0.8\columnwidth]{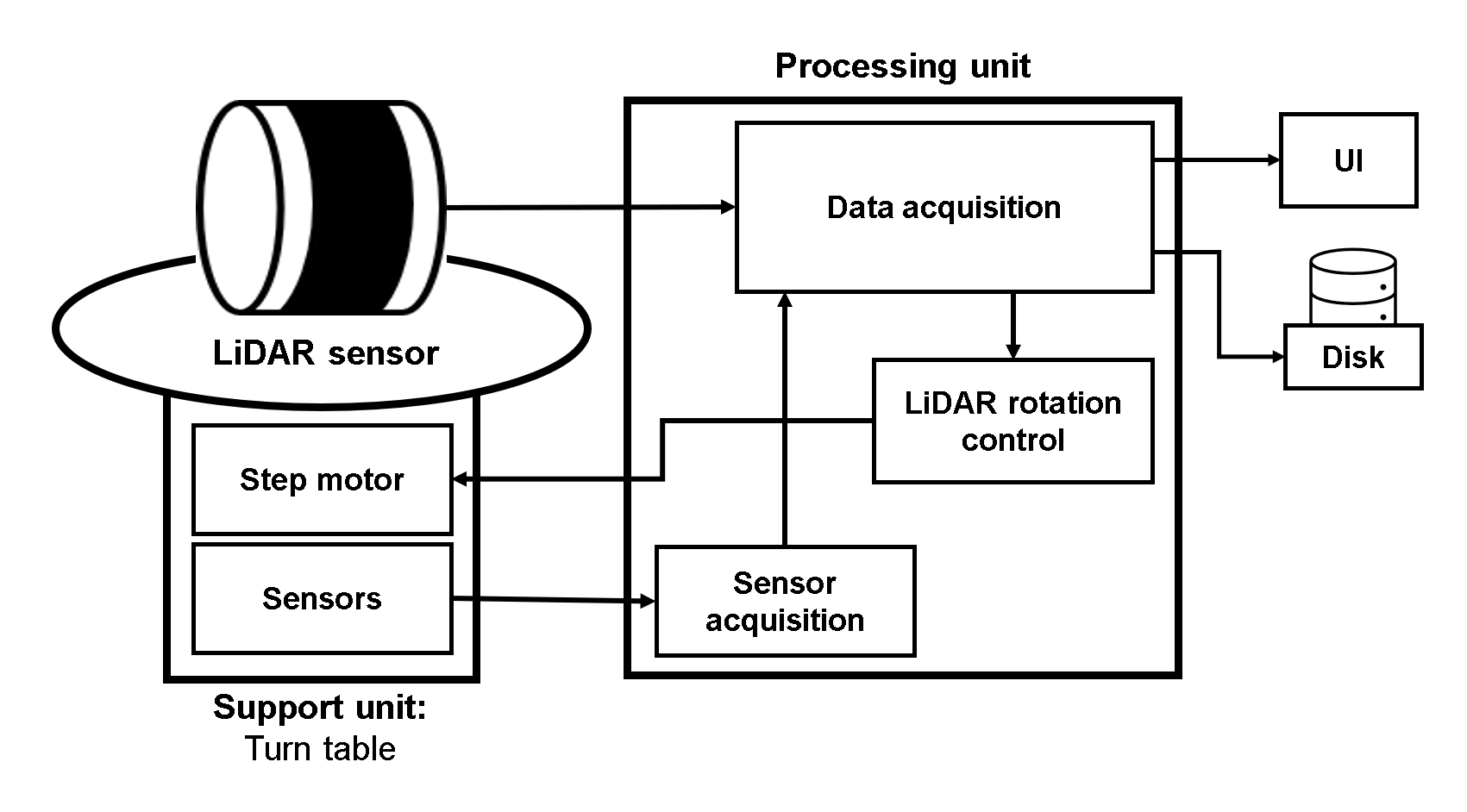}
    \caption{Block diagram of the system with explicit connections between its various components.}
    \label{fig:3.block2}
\end{figure}

Figure \ref{fig:3.block2} shows the relationship between the various components that make the system. The support unit is composed by all hardware components used to complement the LiDAR sensor behaviour and data processing. The processing unit is responsible for all software functionalities, including data acquisition, aggregation and processing. The actuator is mainly composed of motors, used only for LiDAR sensor movement. The sensor block presents examples of available options which can measure location, i.e. Global Positioning System (GPS), or can derive movement information, i.e. gyroscope and accelerometer. Inside the processing unit are various blocks, each with very specific functions. The sensor acquisition block is responsible for acquiring data from all sensors (except LiDAR). Afterwards, this information is sent to data aggregation. This block couples LiDAR data with the previously mentioned sensor data in accordance to the LiDAR position and stores it in disk. During data acquisition, the data processing unit manages the number of samples (number of 360{\textdegree} vertical rotations) required for each vertical plane and creates new instructions to be sent to the motor controller. When data acquisition is finished, the data processing unit is responsible for point cloud computation and send the results to the user interface (UI). Finally, the motor controller is responsible for managing the system movement and rotation of the LiDAR sensor, according the instruction from the processing unit.

\section{Implementation}

\subsection{LiDAR sensor}

\begin{figure} [htb]
\centering
\begin{minipage}{0.45\columnwidth}
\centering
  \includegraphics[width=0.6\textwidth]{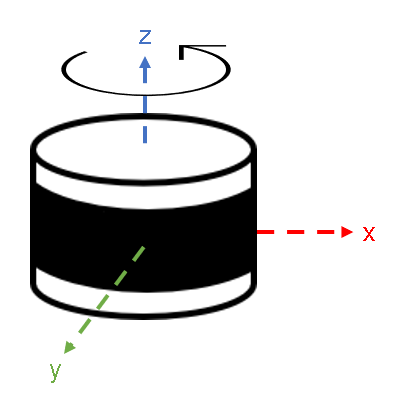}
  
  (a)
\end{minipage}
\begin{minipage}{0.45\columnwidth}
\centering
  \includegraphics[width=0.6\textwidth]{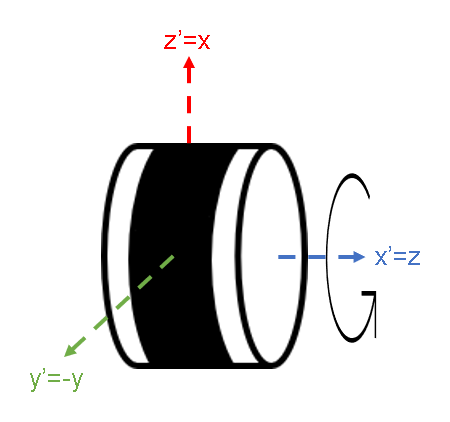}
  
  (b)
\end{minipage}
    
    \caption{(a) Standard orientation of the LiDAR sensor; (b) Proposed orientation of the LiDAR sensor}
    \label{fig:3.sensor}
\end{figure}

The LiDAR sensor model chosen for this system implementation is the VLP-16  model from Velodyne. This sensor makes 16 simultaneous laser emissions on the horizontal plane with an aperture angle of 2{\textdegree}, for a field of view of 360\textdegree x30\textdegree, and up to 2 distance measurements per laser emission, ranging between 0.5m and 100m with an accuracy of 3cm. This sensor is mounted in a non conventional position, in order to capture the vertical plane instead of the typical horizontal plane (Figure \ref{fig:3.sensor}). With this change, the field of view and angular resolution of the sensor are switched. This way, the sensor has a field of view of 30\textdegree x360{\textdegree} and an angular resolution in the horizontal and vertical plane of 2{\textdegree} and between 0.1\textdegree -0.4\textdegree.



{}

\subsection{Hardware implementation for sensor rotation}

\begin{figure} [htb]
    \centering
    \includegraphics[width=0.5\columnwidth]{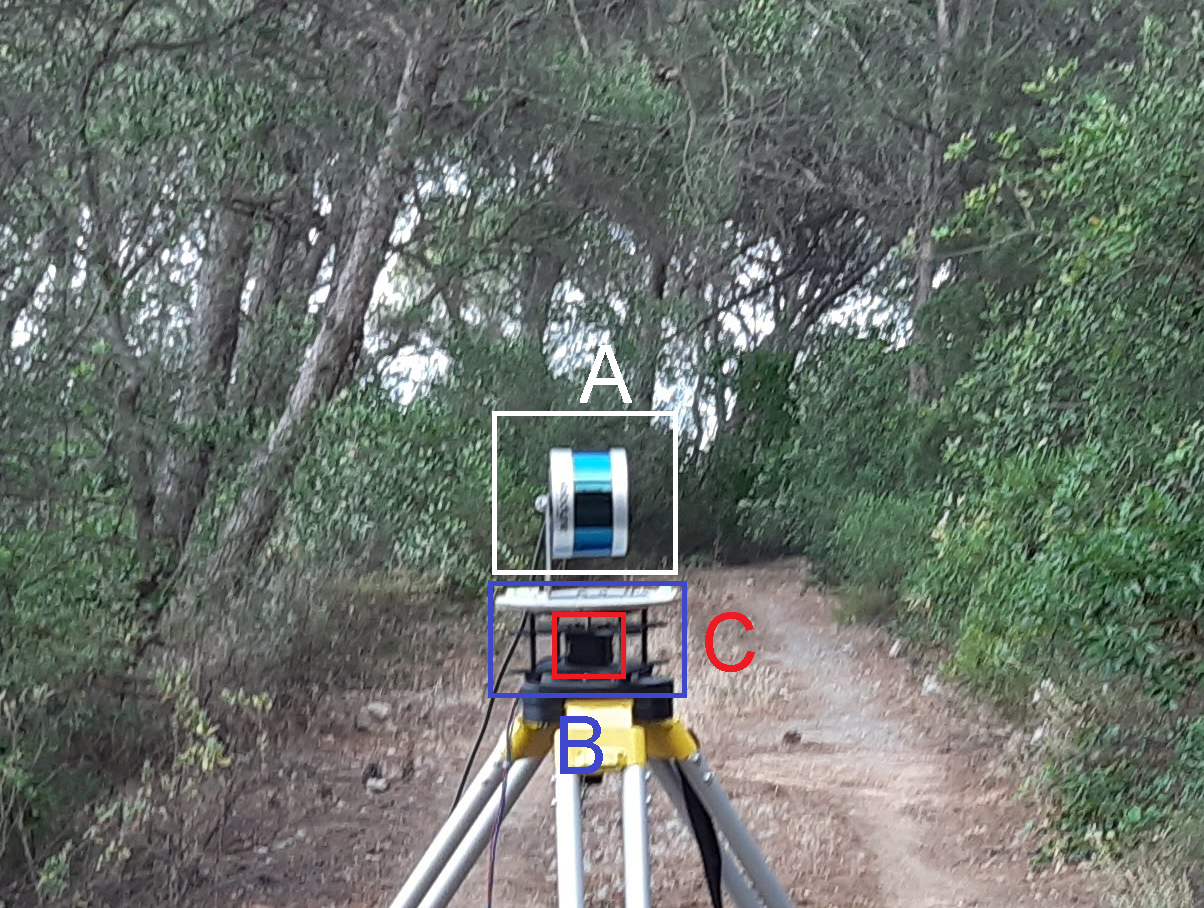}
    \caption{Low Cost stationary LiDAR system. This system is composed by the LiDAR sensor (A), a turntable (B) and a step motor (C).}
    \label{fig:3.system}
\end{figure}

The first objective of this stationary system is to achieve a spherical field of view (360\textdegree x360\textdegree), allowing for an increased data acquisition. To achieve this the LiDAR sensor is mounted on a turn table \cite{turntable}. The step motor used to rotate the table is the Nema 17 model \cite{step_motor}, which has a default step size of 1.8\textdegree. The system implementation using these components is presented on the Figure \ref{fig:3.system}.

Depending on the motor step size, different levels of horizontal angular resolution can be achieved. Since the new orientation of the LiDAR sensor changed its horizontal resolution to 2\textdegree, with a step size of 1.8{\textdegree} and 0.9{\textdegree} the resulting system's horizontal angular resolution value is 0.2{\textdegree} and 0.1{\textdegree}, respectively. Through the configuration of the motor controller it's also possible to further decrease the step size improving horizontal resolution, however, doing so increases the difficulty to validate motor rotation and greatly increases the point cloud capture time.

\subsection{LiDAR rotation control}

\begin{figure}[htb]
    \centering
    \includegraphics[width=0.4\textwidth]{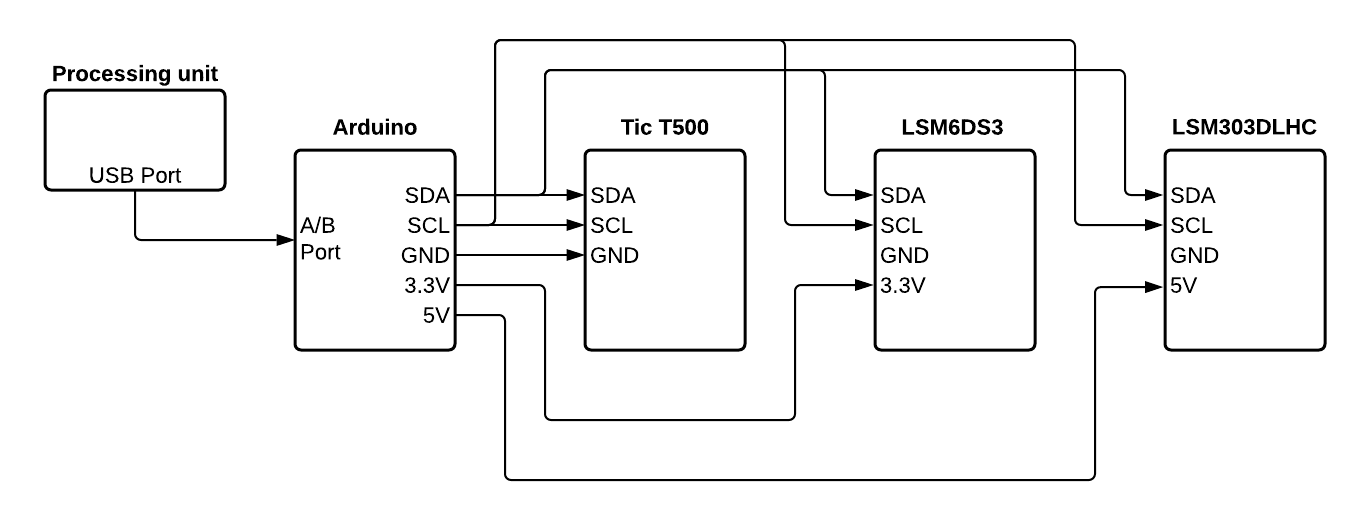}
    \caption{Electric connection between sensors and motor controller.}
    \label{fig:electric}
\end{figure}

The motor is controlled by a combination of a step motor controller and additional sensors. The controller used is the Polulu Tic T500 \cite{controller} model. This controller allows the configuration of motor characteristics, such as rotation speed and acceleration, power consumption and step resolution. Step resolution defines a new step size the motor can perform. The model Tic T500 allows to configure the step size as full, half, 1/4 and 1/8 of the motor's default step size. The sensors used are LSM6DS3 \cite{accel_gyro} (accelerometer and gyroscope) and LSM303DLHC \cite{magn} (magnetometer). The accelerometer and gyroscope are used to complement validation of motor rotation and compute LiDAR tilt and the magnetometer to orient point cloud data to north. Figure \ref{fig:electric} presents the electric connections between these components and with the processing unit.

\subsection{LiDAR tilt}

\begin{figure}[htb]
    \centering
    \scriptsize
    \includegraphics[width=0.25\textwidth]{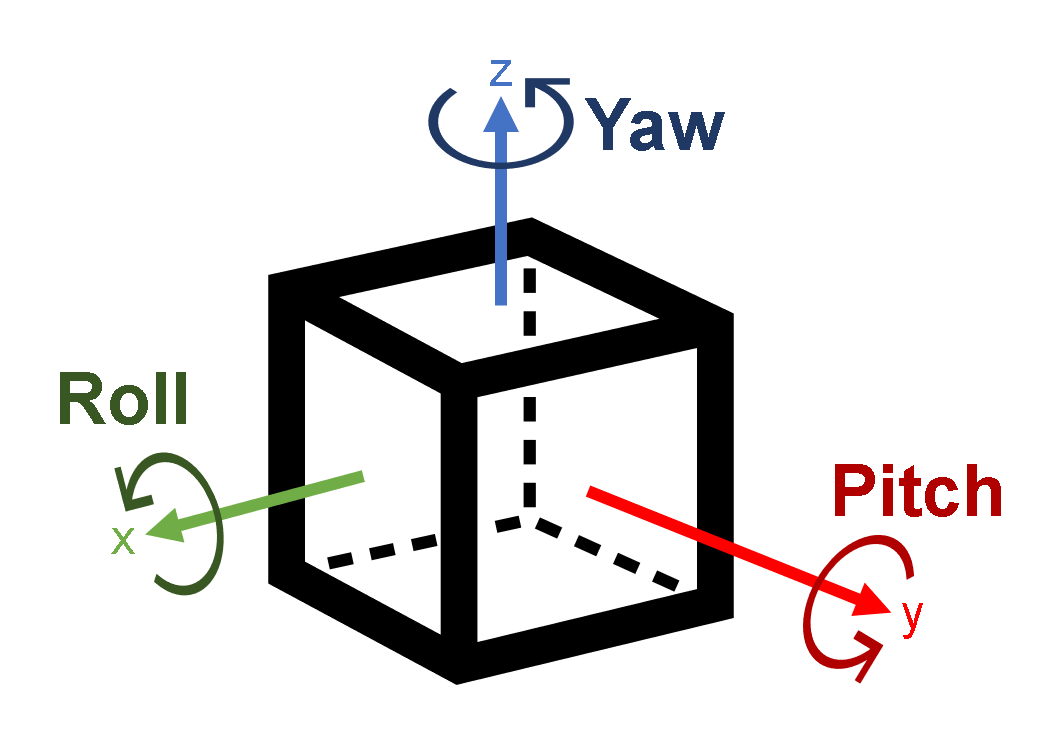}
    \caption{Graphical representation of an object 6 degrees of freedom (x, y and z axis and pitch, roll and yaw angles).}
    \label{fig:3.angles}
\end{figure}

Depending on the studied environment, terrain may not be uniform and wchich results in incorrect environment characterisation. To correct such behaviour, an inertial measurement unit  (IMU) \cite{accel_gyro}, composed of an accelerometer and gyroscope, is employed to automatically compute LiDAR tilt.  Using the measurements of these sensors, roll, pitch and yaw angles (\ref{fig:3.angles}) can be computed using the expression (\ref{eq:3.roll}), (\ref{eq:3.pitch}) and (\ref{eq:3.yaw}), based on the complementary filter implementation from \cite{mccarron2013low}. In these expression, $Roll$, $Pitch$ and $Yaw$ are the desired angles in degrees, $Gx$, $Gy$ and $Gz$ correspond to the gyroscope measurements, $Ax$, $Ay$ and $Az$ correspond to accelerometer measurements, $T$ is the sampling period, including both sensor readings and angle calculations, and $\alpha$ is related to the time constant $\tau$ and the sampling rate through the expression (\ref{eq:3.alpha}). For the implementation of these sensors, the time constant was set to 0.1s.

\begin{equation}
\footnotesize
    Roll = \alpha \times (Roll_{prev} + Gx\times T)  + (1-\alpha)\times arctan\Big(\frac{Ay}{Az} \Big) \\
    \label{eq:3.roll}
\end{equation}

\begin{equation}
\footnotesize
    Pitch = \alpha \times (Pitch_{prev} + Gy\times T)  + (1-\alpha)\times arctan\Big(\frac{-Ax}{\sqrt{Ay^{2}+Az^{2}}} \Big) \\
    \label{eq:3.pitch}
\end{equation}

\begin{equation}
\footnotesize
    Yaw = Yaw_{prev} + Gy\times T\\
    \label{eq:3.yaw}
\end{equation}

\begin{equation}
\footnotesize
    \alpha = \frac{\tau}{\tau + T}\\
    \label{eq:3.alpha}
\end{equation}

\subsection{Data acquisition and processing}

\begin{figure}[htb]
    \centering
    \includegraphics[width=0.25\textwidth]{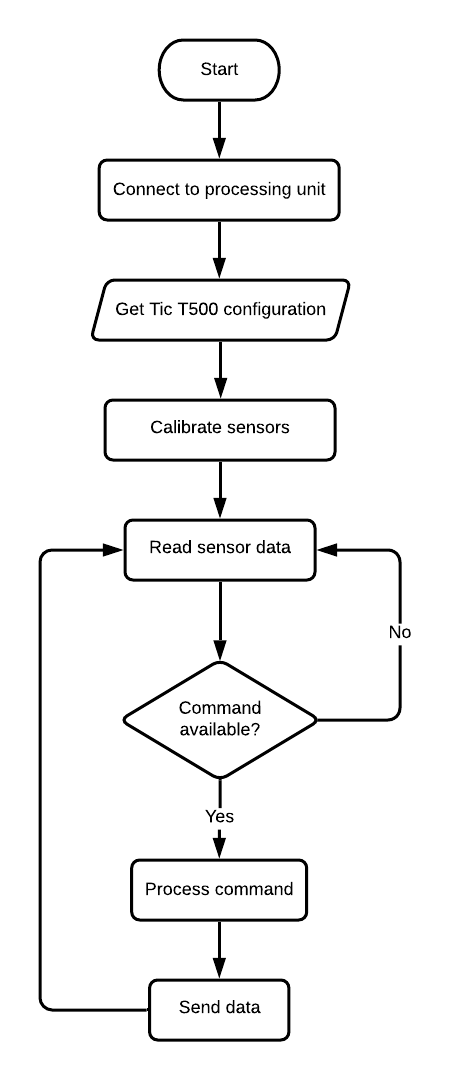}
    \caption{Block diagram representation of the Arduino general initialisation and behaviour.}
    \label{fig:3.arduino}
\end{figure}

Data acquisition is done by two main components, Arduino UNO and a main processing unit. The Arduino manages the behaviour of the motor controller, continuously updates sensor data and also works as a proxy between these components and the processing unit (Figure \ref{fig:3.arduino}).

\begin{figure}[htb]
    \centering
    \includegraphics[width=0.25\textwidth]{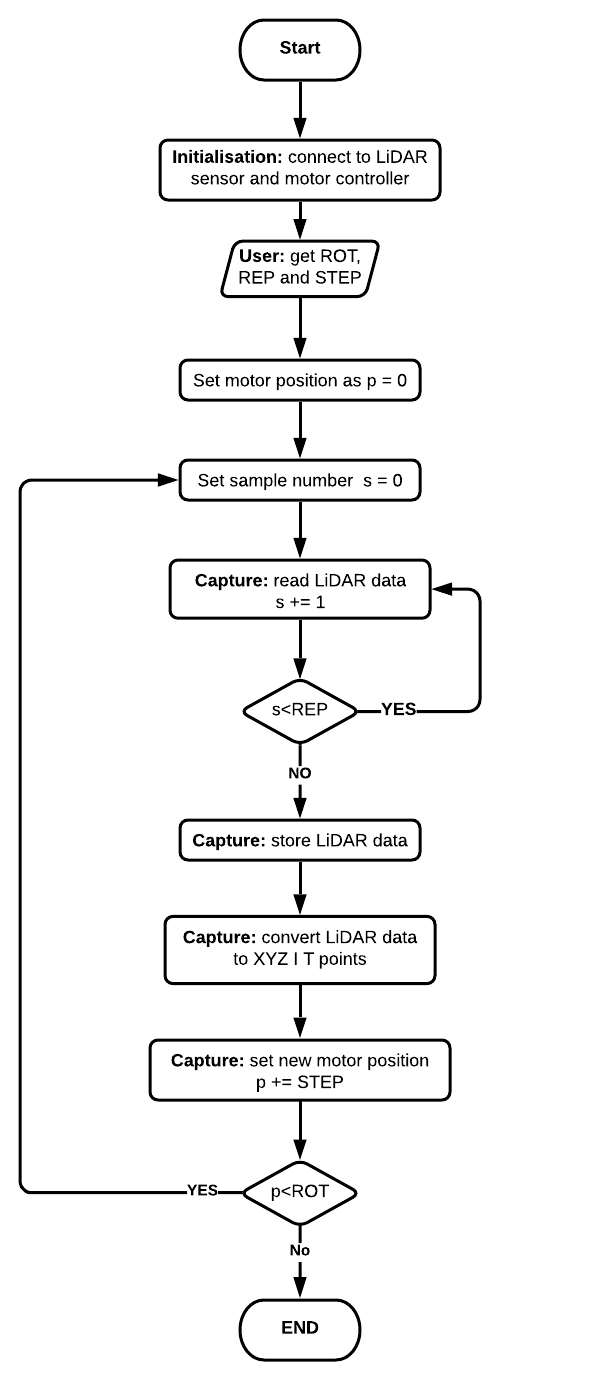}
    \caption{Block diagram representation of a LiDAR point cloud capture workflow. This workflow is specific for a LiDAR stationary system with horizontal rotation performed by a step motor and automatic point cloud correction.}
    \label{fig:workflow}
\end{figure}

The main processing unit reads all LiDAR data, both distance measurements and position provided by GPS, and sensor data from Arduino. Afterwards, XYZ point coordinates are computed from LiDAR data and manipulated using the sensor data. From the block diagram in Figure \ref{fig:workflow}, representing a point cloud capture workflow, to initialise a point cloud capture some user inputs are required: horizontal rotation (ROT), motor step size (STEP) and vertical plane sampling (REP). Horizontal rotation is referred as the maximum LiDAR sensor horizontal rotation and vertical plane sampling as number of samples of complete 360º LiDAR laser rotation per motor rotation. The initial position of the system is set according to the north orientation provided by the compass. The rotation step corresponds to the number of motor steps taken between each vertical plane samplings.


When starting a point cloud capture, the first step is to compute LiDAR tilt. Afterwards, the processing unit reads LiDAR data corresponding to the total number of plane samples desired by the user. The sampled vertical plane is both stored it in a txt file, in its raw format, and converted to XYZ point coordinate, along with its intensity value, I, and timestamp, T (time since the LiDAR sensor has been turned on). GPS is connected directly to the LiDAR sensor. Doing so improves the accuracy of timestamps of each measurement provided. This way, LiDAR becomes responsible of relaying to the processing unit the global position of the system. This data is never stored, since its only used once for point cloud georreferencing and ignored ever since. When data is stored, the processing unit sends new instructions to the actuator control to change to a new position. These two last steps, data reading and motor rotation, are repeated until the LiDAR sensor rotation corresponds to the user defined horizontal rotation. Finally, a point cloud is generated and available for user visualisation and further processing.


\section{LCLS evaluation}

\subsection{Cost}

Table \ref{tab:4.components} presents the main components for the system's implementation, along with their price and corresponding role, according to the architecture presented in the Figure \ref{fig:3.system}. 

\begin{table}[htb]
\captionsetup{font=scriptsize}
\caption{Components used for the LCLS implementation.}
    \centering
    \scriptsize
    \begin{threeparttable}
    \begin{tabular}{l l l}
    Component & Role & Price \\
    \toprule
    Velodyne VLP-16 & LiDAR sensor & 11000€ \\
    Turntable & Support unit & 49.75€ \\
    Tripod & Support unit & 80€ \\
    Nema 17 & Actuator & 15.87€ \\
    Pololu Tic T500 & Actuator controller & 19.62€ \\
    Arduino & Sensor acquisition & 14.98€ \\
    LSM6DS3 & Sensors & 14.27€ \\
    LSM303HLHC & Sensors & 9.59€ \\
    12V baterry & Support unit & 13.95€ \\
    \midrule
    \multicolumn{2}{l}{Total} & 11218.03€\\
    \bottomrule
    \end{tabular}
\end{threeparttable}
    \label{tab:4.components}
\end{table}

The price of the LiDAR sensor VLP-16 model used was 11000€, above the price limit setup as requirements (10000€). This is due to the fact that the LiDAR sensor was bought in the year 2015. Since then, this model has had it's price decreased and many cheaper options with similar characteristics have appeared. The rest of the components proved to be easy to obtain with a total price below 600€, including some miscellaneous components, such as wires and bolts, and component shipment costs. This brings the total cost of the developed system below 11600€ and, since there are many cheaper options for the LiDAR sensor, the total cost can still be reduced.

\subsection{Technical performance}

\subsubsection{Configuration and calibration}


To keep the sampling process uniform, configuration and calibration of the different components used was equal for all study locations. The accelerometer and gyroscope were calibrated by computing their drift value, which corresponds to the mean value of multiple measurements with the sensor stationary in the starting position. The magnetometer measurements suffered no change since the model used comes factory calibrated.





\subsubsection{Distance assessment}

Accuracy and resolution of the distance measurements were evaluated by computing multiple distances between various points. These distances were measured on site with a measurement tape and on the point cloud using CloudCompare \cite{girardeau2016cloudcompare} applications. 

\begin{figure}[htb]
    \centering
    \includegraphics[width=0.25\textwidth]{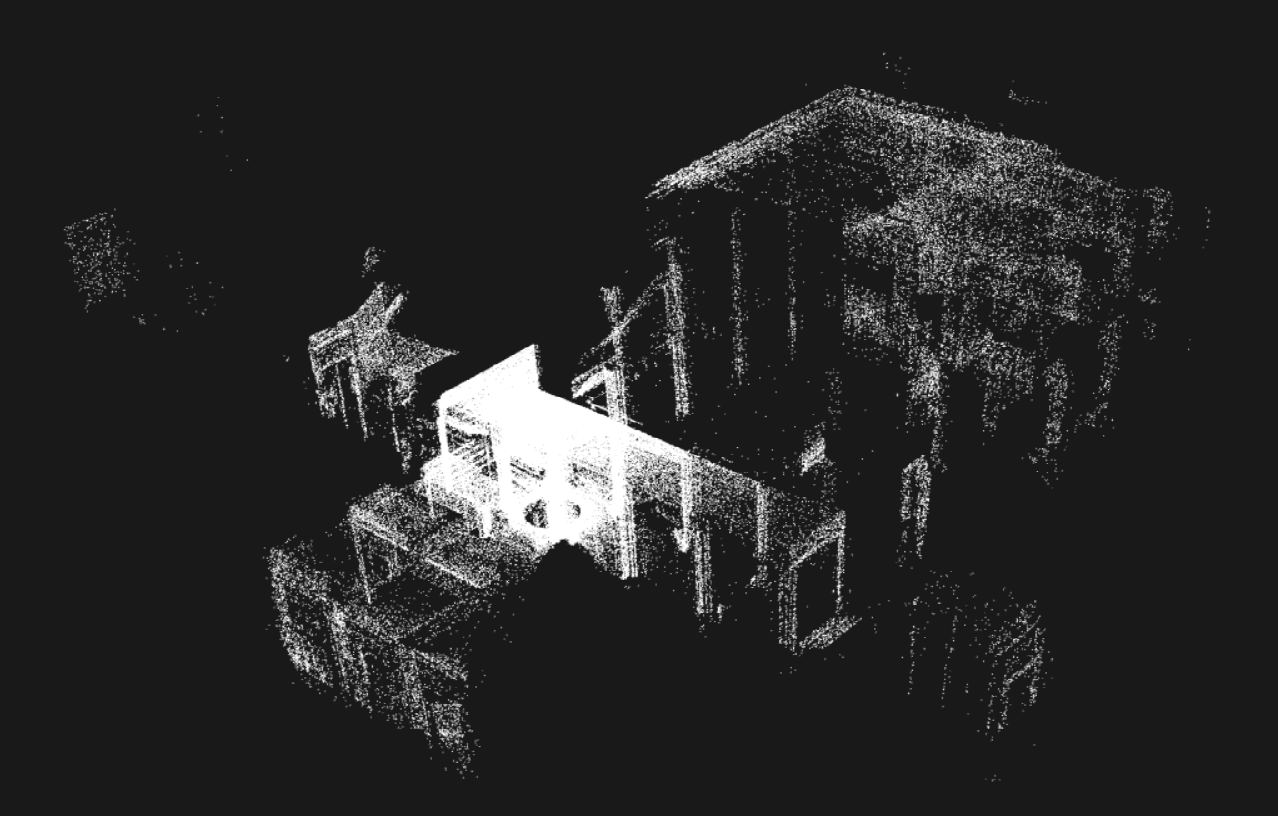}
    \caption{Point cloud data of the central hall at IST.}
    \label{fig:4.ist_pc}
\end{figure}

The study area used to evaluate these characteristics is the main hall of the central pavilion at Instituto Superior Técnico (IST) (38\textdegree44'12.8'' N, 9\textdegree08'21.4'' E), taken on 19 December 2020. Figure \ref{fig:4.ist_pc} presents a visualisation of the acquired point cloud data in this location. 

The resulting measurements, both tape measured and from point cloud data, are presented in the Table \ref{tab:4.acc}. Both H and I are radial lines, beginning close to the LiDAR system and ending at maximum distances of 20-30m away. For this reason, distances from the LiDAR cannot be presented.












\begin{table}[htb]
\captionsetup{font=scriptsize}
\caption{Comparison of distance measurements taken from LiDAR point clouds and manually using a measuring tape.}
    \centering
   \scriptsize
    \begin{threeparttable}
    \begin{tabular}{c c c c c c}
    \multirow{2}{*}{ID} & \multirow{2}{*}{Dist.\tnote{1} [m]} & \multicolumn{2}{c}{Measurements} & \multirow{2}{*}{Error\tnote{2} [m]} & \multirow{2}{*}{Error [\%]}\\
    \cline{3-4}
     &  & Field [m] & LCLS [m] & &\\
    \toprule
    A & 2.11 & 0.78 & 0.75 & -0.03 & -3.85\\
    A & 3.12 & 0.78 & 0.76 & -0.02 & -2.56\\
    A & 6.83 & 0.78 & 0.73 & -0.05 & -6.41\\
    B & 10.35 & 0.64 & 0.68 & 0.04 & 6.25\\
    B & 6.48 & 0.64 & 0.62 & -0.02 & -3.13\\
    B & 2.96 & 0.64 & 0.62 & -0.02 & -3.13\\
    B & 5.80 & 0.64 & 0.62 & -0.02 & -3.13\\
    C & 1.80 & 3.00 & 2.99 & -0.01 & -0.33\\
    C & 3.30 & 3.00 & 2.95 & -0.05 & -1.67\\
    C & 8.90 & 3.00 & 3.00 & 0.00 & 0.00\\
    C & 0.50 & 3.00 & 2.95 & -0.05 & -1.67\\
    D & 5.00 & 2.60 & 2.58 & -0.02 & -0.77\\
    E & -- & 27.86 &28.07 & 0.21 & 0.77\\
    F & -- & 19.37 & 19.07 & -0.3 & -1.55\\
    \bottomrule
    \end{tabular}
    \begin{tablenotes}
    \item[1] Distance to LiDAR sensor
    \item[2] LCLS - Field
    \end{tablenotes}
\end{threeparttable}
    \label{tab:4.acc}
\end{table}

Looking at the results from the Table \ref{tab:4.acc}, distance measurements taken from point cloud were almost always underestimating field measurements. However, the difference between the two types of measurements is around 2cm, which corresponds to the accuracy value of the sensor used. This error is constant along with distance from the sensor. Distances H and I, which correspond to the largest distances measured, had a variation of 0.21m and -0.3m, respectively, from their field measurements, which can be explained by the lack of accuracy when tape measured. 


\begin{table*}[htb]
\captionsetup{font=scriptsize}
\caption{Horizontal resolution values from two different point cloud capture configuration at three distances from the sensor. Measured resolution corresponds to the average value of 10 samples.}
\centering
\scriptsize
\begin{tabular}{l l l l l l l l m{1.5cm}}
\multirow{2}{*}{STEP} & \multirow{2}{*}{REP} & \multirow{2}{*}{ROT} & \multirow{2}{1.7cm}{Angular resolution} & \multirow{2}{*}{Distance} & \multirow{2}{*}{Resolution} & \multicolumn{3}{c}{Measured resolution} \\
\cline{7-9}
 &  &  &  &  & & \multicolumn{2}{c}{Average} & Standard deviation \\
\toprule
\multirow{3}{*}{1} & \multirow{3}{*}{10} & \multirow{3}{*}{180} & \multirow{3}{*}{0.2\textdegree} & 5m & 0.0175m & 0.0416m & 0.477\textdegree & -0.0119 \\
 &  &  &  & 23m & 0.0803m & 0.0887m & 0.221\textdegree & 0.0115 \\
 &  &  &  & 34m & 0.1187m & 0.1185m & 0.200\textdegree & 0.0027 \\
 \midrule
\multirow{3}{*}{2} & \multirow{3}{*}{5} & \multirow{3}{*}{180} & \multirow{3}{*}{0.4\textdegree} & 5m & 0.0349m & 0.0539m & 0.618\textdegree & 0.0124 \\
 &  &  &  & 23m & 0.1606m & 0.1725m & 0.430\textdegree & 0.0280 \\
 &  &  &  & 34m & 0.2373m & 0.2270m & 0.383\textdegree & -0.0316 \\
 \bottomrule
\end{tabular}
\label{tab:4.res}
\end{table*}

Table \ref{tab:4.res} presents the minimum distance between two point cloud points at different distances of the position of the LiDAR sensor. The minimum distance directly provides the resolution value of the system according to the point cloud capture configuration. With a configuration of STEP 1, REP 10 and ROT 180, the resolution varies with the distance to the LiDAR sensor. At short distances (5m) the resolution measured is around 0.042m (0.477\textdegree), higher than the value expected. This deviation is somewhat expected since this value is close to the measuring accuracy from the LiDAR specifications (3cm). At higher distances from the LiDAR (23m and 34m), the measured resolution are similar to the expected value. For a capture configuration with STEP 2, REP 5 and ROT 180, the behaviour of the results are similar to the previous configuration. Together with accuracy results, it's possible to conclude that, for a stationary system, the LiDAR sensor is the main factor which dictates point cloud quality and measurement accuracy. 



\subsection{Position sensors assessment}

\subsubsection{Tilt measurements}

Accuracy of the LiDAR tilt measurements was performed by simulating angle variation and by comparing roll and pitch angles provided by the sensor and manually measured using a protractor (minimum unit of 1\textdegree). In the Table \ref{tab:4.slope} are the measured roll and pitch angles in three different situations: change in roll angle, change in pitch angle and change in both roll and pitch angles. 


Changes in measured and sensed angles are below 3{\textdegree} in all combinations. Furthermore, for simulations of roll and pitch angles, sensor underestimated the measurements obtained with the protractor, which is concluded by the negative error values for roll angle and positive error values for pitch. For the variation of both roll and pitch, the error is positive for roll and negative for pitch, overestimating the protractor measurements. Overall, error variation is within acceptable margins, this way, it can be concluded that the implementation of an accelerometer and gyroscope can be beneficial for the intended use of calculating terrain slope for correction of point cloud data. 


\begin{table}[htb]
\captionsetup{font=scriptsize}
\caption{Sensor angle measurements in three different situations.}
\centering
\scriptsize
\begin{threeparttable}
\begin{tabular}{l l l l l}
 &  & \multicolumn{3}{c}{Measurements} \\
 \cline{3-5}
  \multicolumn{2}{l}{Angles to test} & Field & Sensor & Error\tnote{1}\\
 \toprule
 \multicolumn{2}{l}{\multirow{3}{*}{Roll}} & 14\textdegree & 13.76\textdegree & -0.24\textdegree \\
  & & 24\textdegree & 21.11\textdegree & -2.89\textdegree \\
  & & 37\textdegree & 34.85\textdegree & -2.15\textdegree \\
  \midrule
  \multicolumn{2}{l}{\multirow{3}{*}{Pitch}} & -14\textdegree & -13.35\textdegree & 0.65\textdegree \\
  & & -24\textdegree & -21.90\textdegree & 2.01\textdegree \\
  & & -32\textdegree & -29.61\textdegree & 2.39\textdegree \\
  \midrule
  \multirow{6}{*}{Roll and pitch} & \multirow{3}{*}{Roll} & 23\textdegree & 20.44\textdegree & -2.56\textdegree\\
  & & 15\textdegree & 16.58\textdegree & 1.58\textdegree \\
  & & 31\textdegree & 31.85\textdegree & 0.85\textdegree \\
  \cline{2-5}
  & \multirow{3}{*}{Pitch} & -11\textdegree & -11.30\textdegree & -0.3\textdegree \\
  & & -25\textdegree & -25.33\textdegree & -0.33\textdegree \\
  & & -20\textdegree & -21.92\textdegree & -1.98\textdegree \\
 \bottomrule
\end{tabular}
\begin{tablenotes}
\item[1] Error calculated as Sensor - Field.
\end{tablenotes}
\end{threeparttable}
\label{tab:4.slope}
\end{table}

\subsubsection{Compass heading}

The system's compass was implemented using a magnetometer, which provides a clockwise angle rotation to north. To test it's accuracy, the angle provided by the magnetometer was compared to measured angles provided by compass a application of a cellphone. Three angle measurements were tested, at 0{\textdegree}, 130{\textdegree} and 280{\textdegree} angles taken with the cellphone application. Results are shown in the Table \ref{tab:4.compass} and indicate error increase with azimuth, with a minimum of 3.5{\textdegree}, when north-oriented, an a maximum of 21{\textdegree} at 130{\textdegree},corresponding to a 16\% error. This deviation between methods could have originated due to calibration from both the cellphone application and factory calibration of the magnetometer. Based on these preliminary results, no conclusion can be taken from the performance of this sensor for calculating angle difference with magnetic north and that further research is needed to account for these differences. A possible replacement for this sensor is to use compass heading computation methods based on gravity measurements \cite{manos2019gravity}, which can be acquired through accelerometers and gyroscopes.

\begin{table}[htb]
\captionsetup{font=scriptsize}
\caption{Measured angle difference with north.}
\centering
\scriptsize
\begin{threeparttable}
\begin{tabular}{l l l l}
 Cellphone & Sensor & \multicolumn{2}{c}{Error\tnote{1}} \\
 \toprule
 280\textdegree & 314.19\textdegree & 34.19\textdegree& 12.21\% \\
 0{\textdegree} (360\textdegree) & 356.5\textdegree & -3.5\textdegree & -0.97\%\\
 130\textdegree & 109.01\textdegree & -20.99 \textdegree & -16.15\%\\
 \bottomrule
\end{tabular}
\begin{tablenotes}
\item[1] Error calculated as Sensor - Manual.
\end{tablenotes}
\end{threeparttable}
\label{tab:4.compass}
\end{table}

\subsection{Execution time and point cloud quality}


Configuration of the LiDAR capture not only dictates the maximum horizontal resolution but also maximum number of points in the resulting point cloud and point cloud acquisition time. To evaluate these characteristics, multiple point clouds were generated with different capture configurations and execution time was measured. Number of points can be calculated using the expression (\ref{eq:4.num_points}), where $NP$ is the number of points, $STEP$, $REP$ and $ROT$ correspond to the user required configuration, $mtSTEP$ is the step motor size in degrees and $vData$ is the number of points acquired per rotation. For VLP-16, the LiDAR sensor model used for the LCLS implementation, to acquire a 360{\textdegree} vertical plane, 82 UDP packets are needed and each provides 32 distance measurements, as such, $vData$ value for this model is $82\times 32= 2624$. The number of points calculated using this expression is equal to the number of points generated during point cloud capture.

\begin{equation}
\footnotesize
    NP = \frac{ROT}{STEP\times mtSTEP}\times REP\times vData
    \label{eq:4.num_points}
\end{equation}

Looking at the results in the Table \ref{tab:4.time}, execution time of acquisitions taken with ROT equal to 360{\textdegree} are double of acquisition taken with ROT equal to 180{\textdegree}. Increasing REP value also greatly increases the amount of points, dictated as number of points (REP=a) = a $\times$ number of points (REP=1), with very little increase in execution time, i.e. around 30s per REP value for STEP of 1 and ROT of 180{\textdegree}. Additionally, some configurations can be more efficient in point cloud acquisition than others. This can be observed by comparing the configurations REP 10 and ROT 180{\textdegree} with REP 5 and ROT 360{\textdegree}, using any STEP value for both configurations. Both configuration result in the same total number of acquired points and the former (REP 10 ROT 180{\textdegree}) has an execution time around half of the latter (REP 5 ROT 360{\textdegree}). This way, it's suggested that the ROT value is maintained at 180{\textdegree}, change STEP value according to the desired horizontal resolution and use REP to define the amount of points and detail of vertical planes of the resulting point cloud.

\begin{table}[htb]
\captionsetup{font=scriptsize}
\caption{Execution time and number of effective points acquired for various types of capture parameters.}
    \centering
    \scriptsize

    \begin{tabular}{l l l l l}
    STEP & REP & ROT & Execution time [min:s] & Number of points\\
    \toprule
    \multirow{6}{*}{1} & 1 & \multirow{3}{*}{180} & 8:46 & 262400\\
    & 5 & & 10:19 & 1312000\\
    & 10 & & 11:45 & 2624000\\
    \cline{2-5}
    & 1 & \multirow{3}{*}{360} & 17:27 & 524800\\
    & 5 & & 20:07 & 2624000\\
    & 10 & & 23:26 & 5248000\\
    \midrule
    \multirow{6}{*}{5} & 1 & \multirow{3}{*}{180} & 1:49 & 52480\\
    & 5 & & 2:05 & 262400\\
    & 10 & & 2:26 & 524800\\
    \cline{2-5}
    & 1 & \multirow{3}{*}{360} & 3:39 & 104960 \\
    & 5 & & 4:10 & 524800\\
    & 10 & & 4:53 & 1049600 \\
    \bottomrule
    \end{tabular}
    \label{tab:4.time}

\end{table}


\subsection{Measuring vegetation complexity}

\subsubsection{Case studies}

\begin{figure} [htb]
    \centering
    \includegraphics[width=0.45\textwidth]{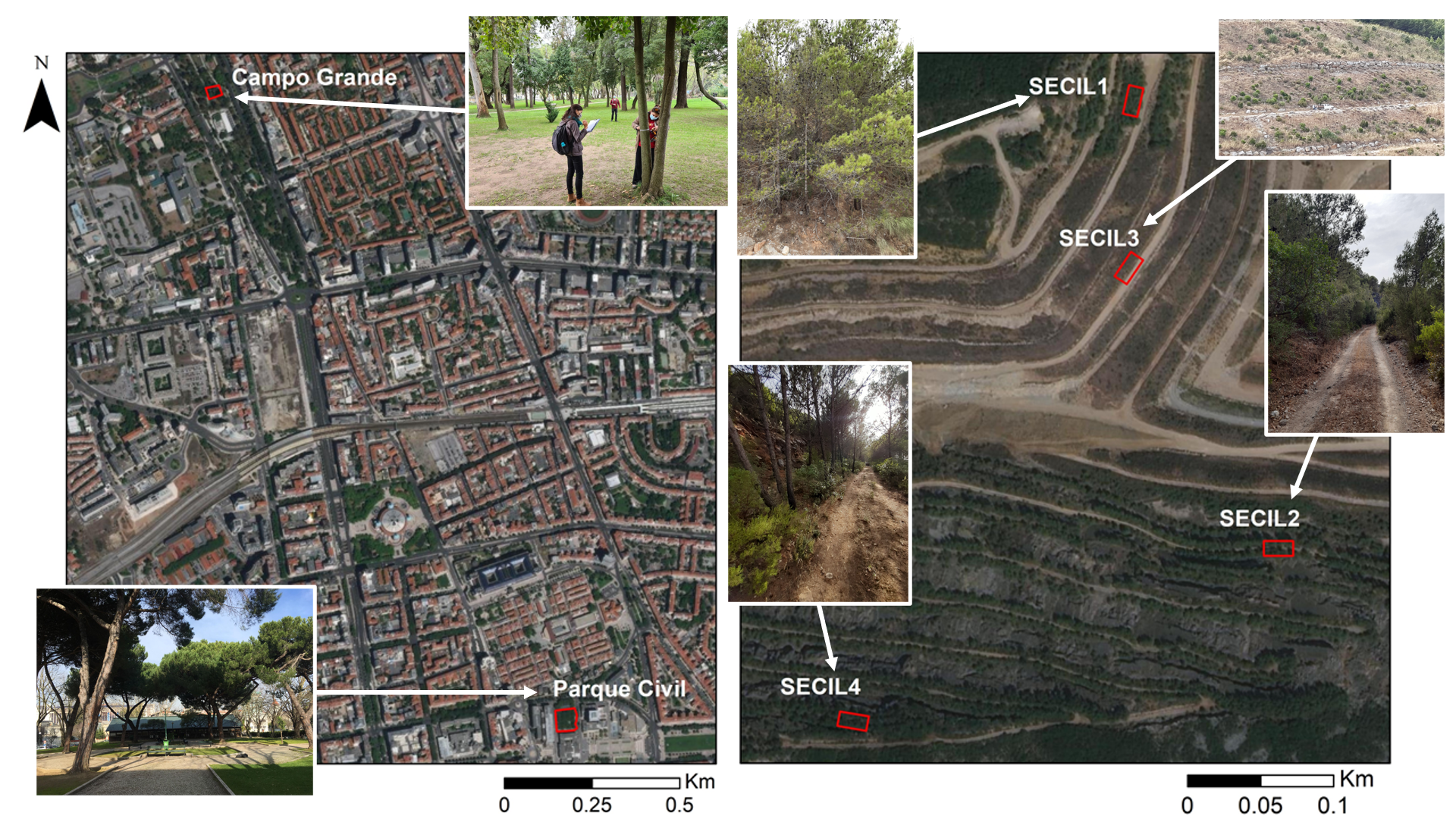}
    \caption{Location map of the sampled urban gardens and the different plots at the limestone quarry.}
    \label{fig:4.map}
\end{figure}

The developed LCLS was used to capture point cloud data from two locations with very distinct types of vegetation structure. A first set of sampling were made at two urban gardens, at the park in Campo Grande (CG) (38\textdegree45'12.7'' N, 9\textdegree09'04.0'' W) and a garden at IST (PC) (38\textdegree44'15.4'' N, 9\textdegree08'22.2'' W). Both of these sites have similar type of vegetation, consisting of a small number of shrubs, which are often pruned, and trees that have similar shape and height. The second set of sampling was done in a limestone quarry (SECIL) (38\textdegree29'43.0'' N, 8\textdegree57'02.5'' W), where LiDAR captures were taken on four locations under ecological restoration, each with different age. Figure \ref{fig:4.map} shows the location of all sampling locations. 


\subsubsection{Data acquisition}


Field sampling at the urban gardens in CG and PC consisted of measuring tree height and diameter at breast height DBH. For the sites at SECIL, a limestone quarry, shrub cover and height and tree density, height and DBH were measured by a research team from the Centre for Ecology, Evolution and Environmental Changes of the University of Lisbon. Shrub cover and height values were obtained by using a 2x10m line intercept method within a plot size of 10x10m. This method consists in extending a tape to create a transect across the site. The observer would then identify all plants intercepted by the tape and record each intercept distance, maximum height, and width. Cover is calculated by adding all measured shrub width and expressing this total as proportion of tape length \cite{canfield1941application}. Tree height and DBH were measured within a plot size of 10x20m. Tree height was measured using a Nikon Forestry Pro laser rangefinder. 

For point cloud captures, the LCLS was mounted on a tripod and settled near the centre of each plot, whenever possible, with consideration of occlusion effects in order to increase environment characterisation within sampling plot. The configuration used for point cloud capture was STEP 1, REP 10 and ROT 180. Points clouds of the SECIL locations were taken on 20 of July 2020, PC point cloud was taken on 22 of July 2020 and CG was taken on 28 of July 2020. 

CG and PC urban gardens share some characteristics, as they have low tree density, of 0.02 trees/m${^2}$ and 0.006 trees/m${^2}$, respectively. At these study sites, tree are quite tall, with maximum heights of 33.4m and 20.4m (Table \ref{tab:4.sites}). The plots at the limestone quarry are all different from the urban gardens, and also between themselves. SECIL2 and 4 develop on horizontal quarry platforms and are interrupted by a road that allows access to those revegetated areas of the quarry. SECIL 1 and 3 correspond to sloped surfaces developing on the side of the road. In these two cases, the LiDAR sensor could not be positioned in the centre of the plot, but was instead placed in the middle of its lower long edge. SECIL 1, 2 and 4 have high tree density of 0.19 trees/m${^2}$, 0.15 trees/m${^2}$ and 0.09 trees/m${^2}$. Tree size between these plots show averages of 8.5m, 10.8m and 12m. Furthermore, these plots are quite different in the shrub stratum development, as can be seen in shrub average height and total cover (Table \ref{tab:4.sites}). SECIL3 has no trees and only contains shrubes.

\begin{figure} [htb]
\centering
  \includegraphics[width=0.25\textwidth]{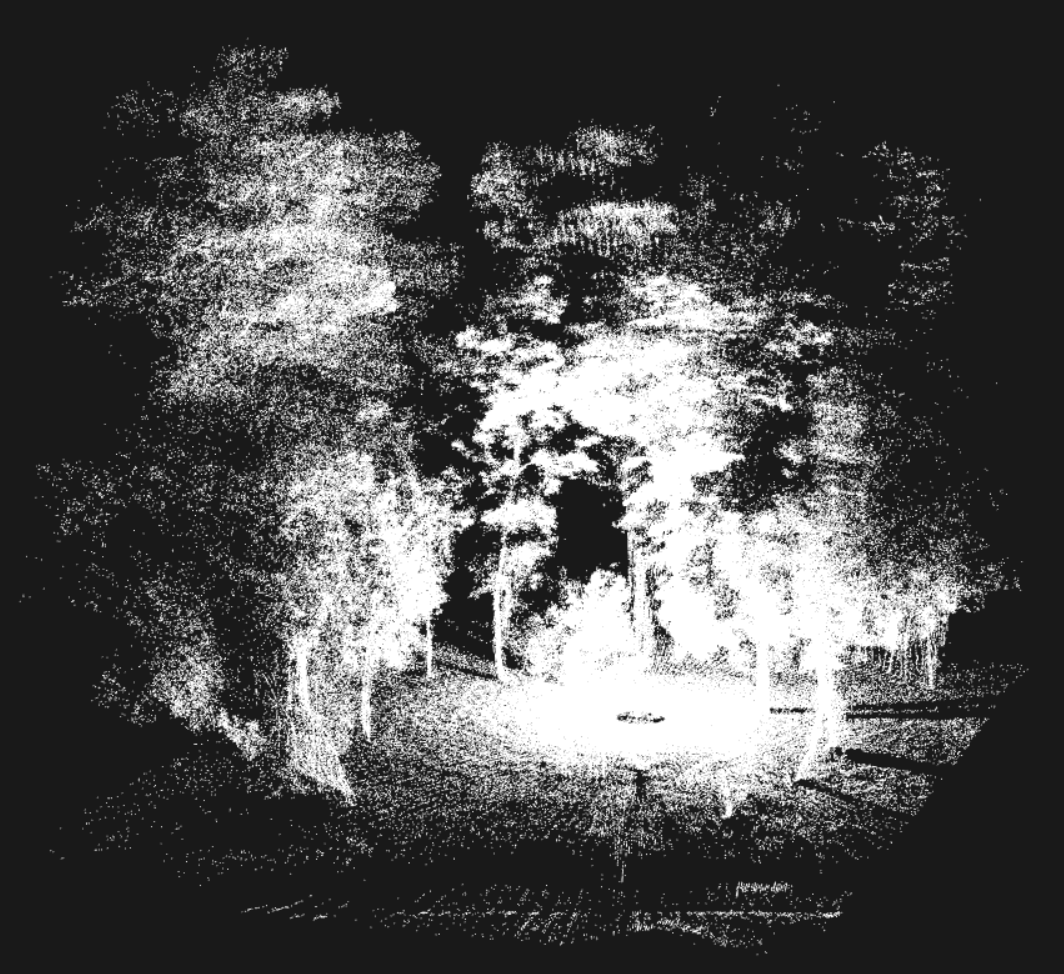}

     \caption{Point cloud visualisation of the studied plot at CG.}
    \label{fig:cg}
\end{figure}

\begin{figure} [htb]

\centering
  \includegraphics[width=0.25\textwidth]{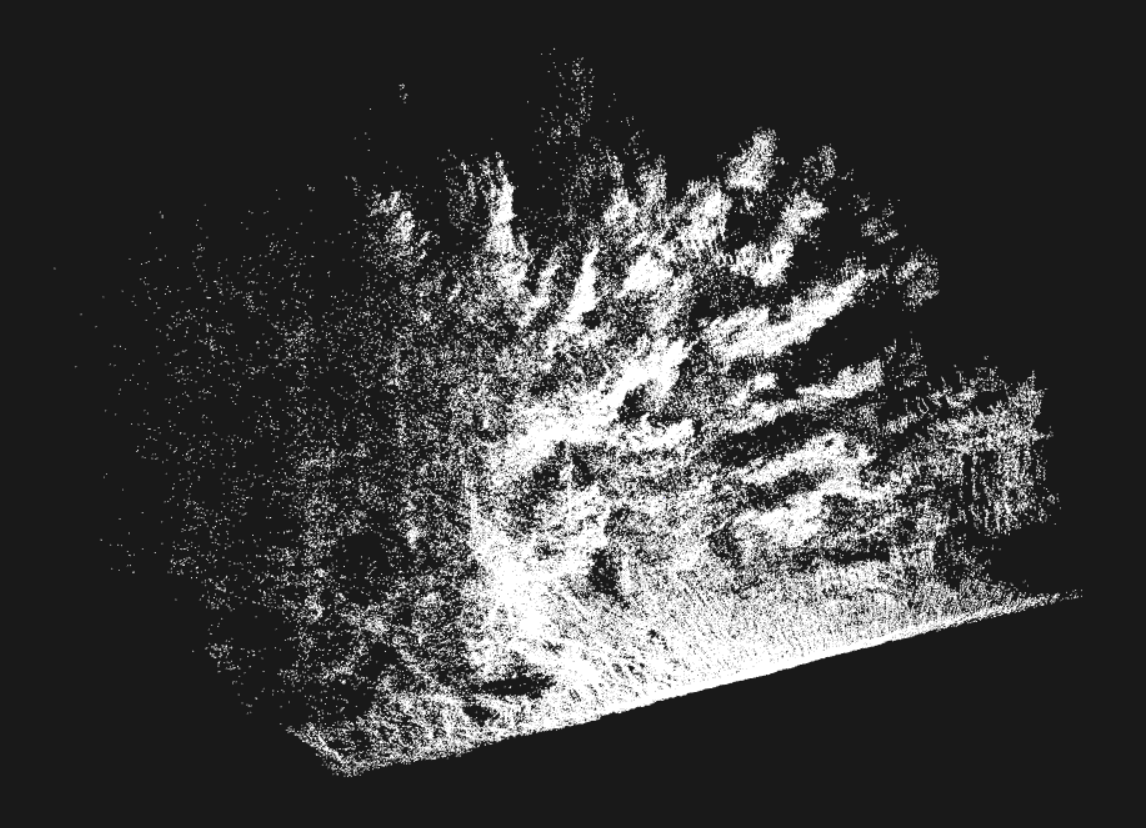}

    \caption{Point cloud visualisation of the studied plot at SECIL1.}
    \label{fig:secil1}
\end{figure}

\begin{figure} [htb]

\centering
 \includegraphics[width=0.25\textwidth]{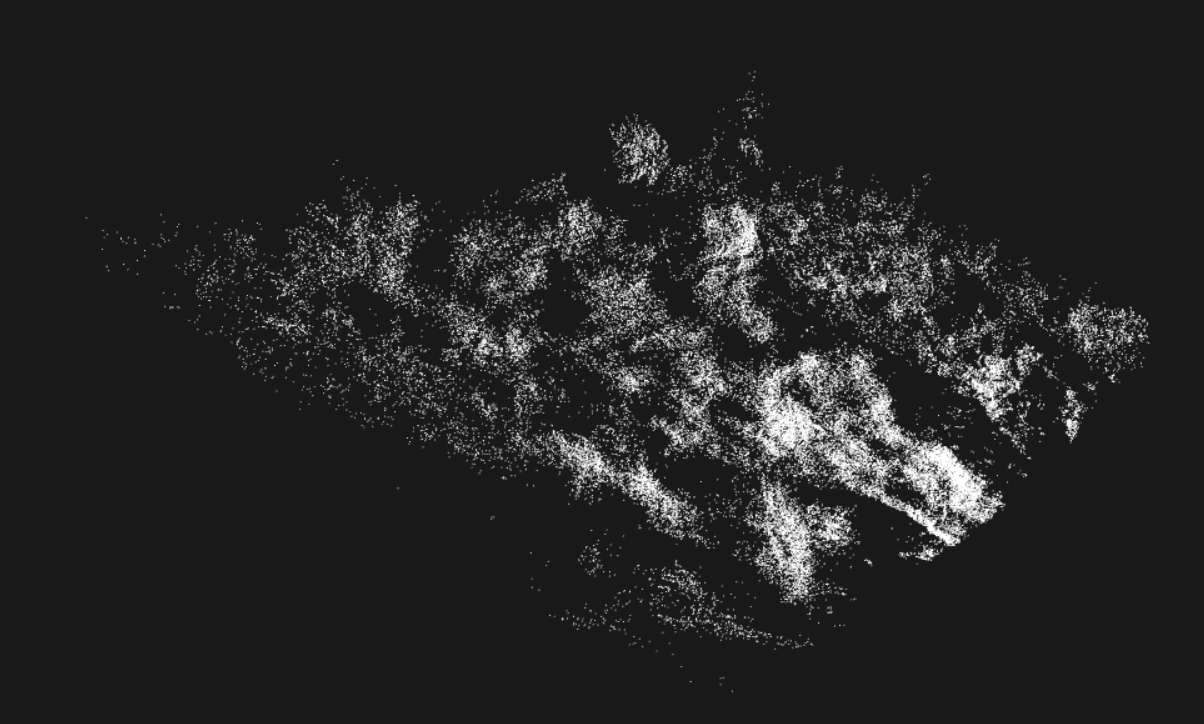}

    \caption{Point cloud visualisation of the studied plot at SECIL3.}
    \label{fig:secil3}
\end{figure}

The Figures \ref{fig:cg}, \ref{fig:secil1} and \ref{fig:secil3} present the resulting visualisation of some of the acquired point clouds. These images were taken using visualisation functions from Python-PCL \cite{python-pcl} library, a Point Cloud Library (PCL), presented in chapter 2, API interface compatible with the software programming language Python.

\begin{table*}[htb]
\captionsetup{font=scriptsize}
\caption{Tree and shrub characterisation.}
    \centering
   \scriptsize
    \begin{tabular}{m{1.5cm}m{1cm}m{2cm}m{1.55cm}m{1.5cm}|m{2.4cm}m{1.8cm}}
    \multirow{2}{1cm}{Case study} & \multirow{2}{1cm}{Plot area [m$^2$]} & \multicolumn{3}{c}{Tree measurement} & \multicolumn{2}{c}{Shrub measurements} \\
    & & Tree density [trees/m$^2$] & Max. tree height [m] & Mean DBH [cm] & Average shrub height [m] & Shrub total cover [\%] \\
    \toprule
    PC & 3529 & 0.006 & 20.4 & 51.7 & 1.5 & 3 \\
    CG & 1168 & 0.02 & 33.4 & 44.6 & -- & -- \\
    SECIL1 & 200 & 0.19 & 8.5 & 12.7 & 100 & 27 \\
    SECIL2 & 200 & 0.18 & 10.8 & 14.8 & 170 & 67 \\
    SECIL3 & 200 & -- & -- & -- & 60 & 44 \\
    SECIL4 & 200 & 0.09 & 12.0 & 23.78 & 58 & 71 \\
    \bottomrule
    \end{tabular}
    \label{tab:4.sites}
\end{table*}



{}

{}


  

{}




\subsubsection{Histograms of vegetation structure}



\begin{figure} [htb]
    \centering
    \includegraphics[width=0.45\textwidth]{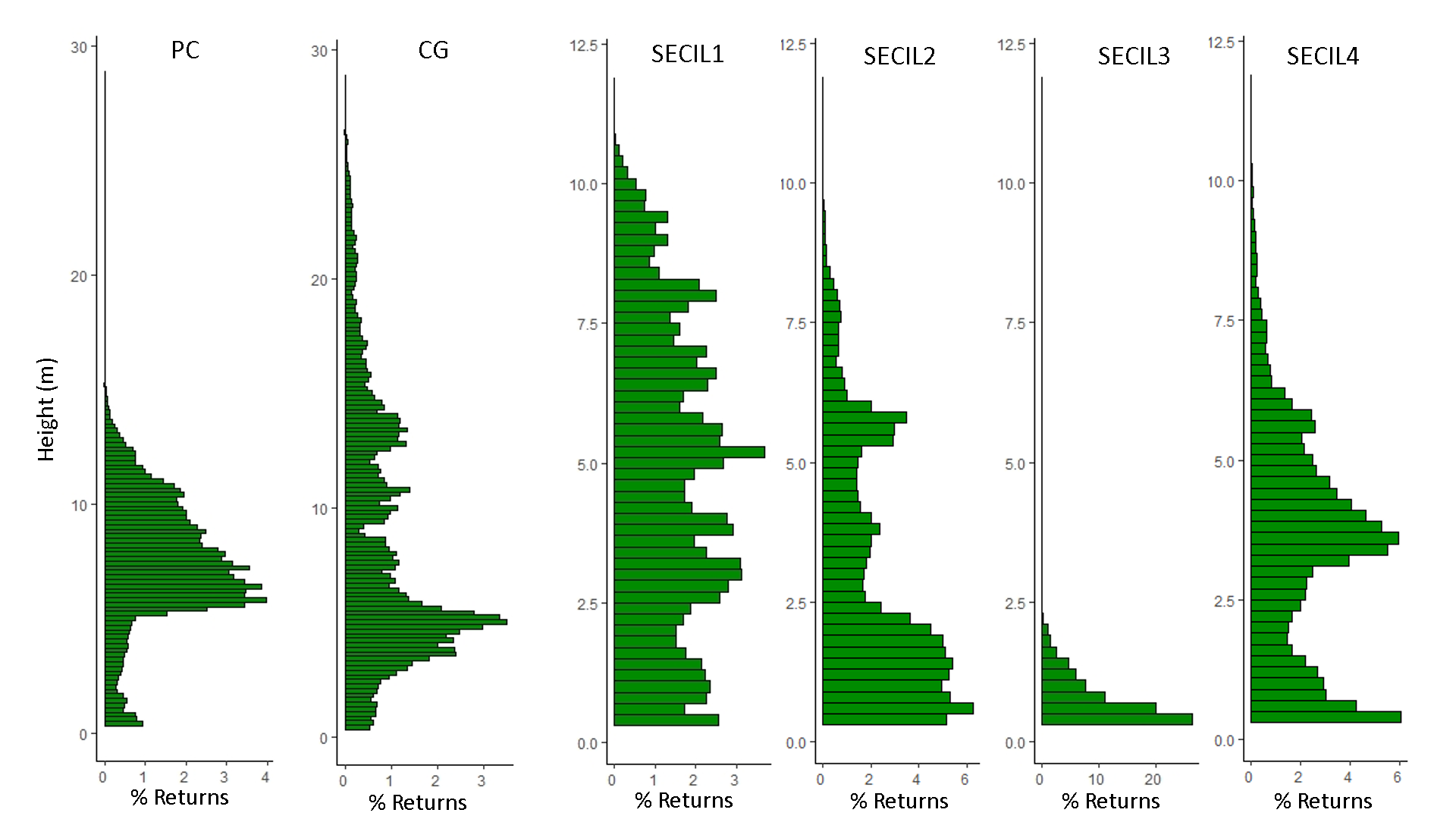}
    \caption{Histogram of vegetation structure for each studied location.}
    \label{fig:4.histogram}
\end{figure}

Histograms of vegetation structure represent the variation of the number of points according to their height to ground, which are a proxy for leaf density. The first step to create these histograms is to filter all duplicate points, which can be done using the function filter\_ duplicate from lidR package \cite{roussel2018lidr, ROUSSEL2020112061}, a R package for ALS data processing. This function excludes points with identical XYZ coordinates, as they represent the same object and can give false idea of changes in leaf density. Height to ground of each points can be calculated using an overlapping digital terrain model (DTM) of the studied area and the function normalize\_ height, also from the lidR package. DTM of each location was computed using the tool las2dem from the LAStools \cite{hug2004advanced} application. The calculated heights can then be represented as a histogram using the function ggplot, from the ggplot2 \cite{wickham2011ggplot2} R package. Figure \ref{fig:4.histogram} presents these results for each location. These histograms represent number of points as percentage of total number of points in height intervals of 20cm, starting at 20cm above ground. The interval 0cm-20cm is excluded since it holds the largest amount of points due to ground returns, which would decrease visualisation of the other height ranges, corresponding to the actual vegetation.

The resulting histograms of vegetation clearly show differences observed in field measurements. Looking at the vegetation structure histograms (figure \ref{fig:4.histogram}), point cloud data allowed to distinguish two main vegetation structures of the environment, the shrub and tree layers, separated by a minimum percentage of returns, which represent an horizontal layer mostly containing three trunks and showing minimum leaf density. Both urban gardens show highest trees with maximum values surpassing 15m in PC and 25m in CG. Furthermore, its also possible to confirm that returns are minimum for lower heights, reflecting a poorly developed or absent shrub stratum, when compared to that of trees. In what concerns the SECIL quarry plots, the absence of tree stratum in SECIL3 is clearly visible in the absent number of returns at heights above 2.5m. In addition, the large size of shrubes in SECIL2 is also visible in the percentage of returns at heights lower than 2.5m. The highest percentage of returns above the 2.5m height in SECIL4 reflects the largest tree size (and consequently canopy size and leaf density), reflected in the largest average DBH measured in the field (Table \ref{tab:4.sites}).

\subsubsection{Vegetation structural index}

The diversity of vegetation height classification was calculated using the Shannon index, which is the most common diversity index applied to heights in LiDAR studies in ecology \cite{listopad2015structural, listopad2018effect, fisher2014savanna}, which numerically represent the vegetation complexity in the vertical stratum for specific class of heights (here, 50cm height classes were used). The Shannon index was determined using the diversity function from vegan \cite{oksanen2016vegan}, a R package which provides tools for basic diversity analyses, community ordination and dissimilarity analysis. 



\begin{table}[htb]

\captionsetup{font=scriptsize}
\caption{Shannon index for each sampled location.}
    \centering
    \scriptsize
    \begin{tabular}{l l}
    Case studies & Shannon index \\
    \toprule
    PC & 3.04 \\
    CG & 3.61\\
    SECIL1 & 2.95\\
    SECIL2 & 2.59 \\
    SECIL3 & 0.94 \\
    SECIL4 & 2.62 \\
    \bottomrule
    \end{tabular}
    \label{tab:4.index}
\end{table}

Shannon diversity index is a quantitative measurement that reflects the diversity of vegetation with different height structure, where taller vegetation structures will have larger height diversity. Amongst the urban parks, CG has the highest vegetation structural diversity than PC. Between SECIL plots, SECIL1 shows highest structural diversity and SECIL3 is the opposite, showing lower diversity. Overall, the shrub cover detected in the field is comparable with the proportion of shrubs returns calculated with the LiDAR system.

\subsubsection{Point cloud data comparison with field data}

To serve as comparison between point cloud data and field data, for each location, maximum height, shrub total cover, height cut-off and returns were determined based on point cloud data. Shrub height cut-off corresponds to the estimated maximum height of shrubs retrieved from the vegetation height histograms, which also corresponds to a minimum percentage return in middle of the histogram, separating shrub from tree stratum (figure \ref{fig:4.histogram}). Shrub return is an estimation of shrub cover calculated from point cloud data. This estimation is calculated using expression \ref{eq:4.shrub_return}, where NSR is the number of shrub returns, NSP the number of shrub points and NGP is the number of ground points. Table \ref{tab:4.shrub} presents these measurements for each locations.

\begin{equation}
    NSR = \frac{NSP}{NSP + NGP}
    \label{eq:4.shrub_return}
\end{equation}

\begin{table}[htb]
\captionsetup{font=scriptsize}
\caption{Environment characterisation using point cloud data. SHCO is shrub height cut-off, NGP is the number of ground points, NSP is the number of shrub points, NSR is the number of shrub returns.}
    \centering
   \scriptsize
    \begin{tabular}{c c c c c c}
     Plot & Height [m] & SHCO [m] & NGP & NSP & NSR [$\%$] \\
    \toprule
    PC & 17.09 & 1.5 & 394638 & 41410 & 1 \\
    CG & 28.01 & -- & -- & -- & -- \\
    SECIL1 & 11.34 & 2 & 34622 & 138194 & 28 \\
    SECIL2 & 10.75 & 2.5 & 290198 & 400656 & 58 \\
    SECIL3 & -- & 2.5 & 104003 & 45337 & 43.60 \\
    SECIL4 & 11.47 & 2 & 224888 & 217006 & 49 \\
    \bottomrule
    \end{tabular}
    \label{tab:4.shrub}
\end{table}

For the urban gardens, PC and CG, maximum tree height in Table \ref{tab:4.shrub} are greatly underestimated those from \ref{tab:4.sites}, up to difference of 5.39m,  while for the locations at SECIL, although also underestimated, have a difference up to 0.53m. This variation is due to vegetation density of tree canopy, which can produce great level of occlusion. 

Summarising, field observations agree well with graphical representations of the point cloud (figure \ref{fig:4.histogram}) and with the structural diversity index applied. This shows that the developed LiDAR system can detect changes in the vegetation structure, even in areas with high vegetation density.


\subsection{MLS system comparison}

For further assessment, the acquired data from the developed LiDAR system was also compared with data acquired by a VUX LiDAR system. VUX is a MLS type system, developed by Albatroz Engenharia SA. \cite{albatroz}, which performs data acquisitions in the vertical plane along an oblique line while in constant movement for point acquisition of great areas. The sensor used in this implementation is RIEGL's VUX-1UAV \cite{riegl_vux}, a waveform LiDAR designed to meet measurement performance and integration requirements of ALS. This sensor provides high speed data acquisition using solely one narrow laser beam, compromising to a field of view of 330{\textdegree} over one plane, with a maximum range of 1050 m, dependent on laser pulse repetition rate. This sensor was implemented in order to acquire oblique lines in the vertical plane. This method allows to change the observation method of vegetation, increasing its characterisation.

\subsubsection{Case studies}

\begin{figure} [htb]
\centering
    \includegraphics[width=0.25\textwidth]{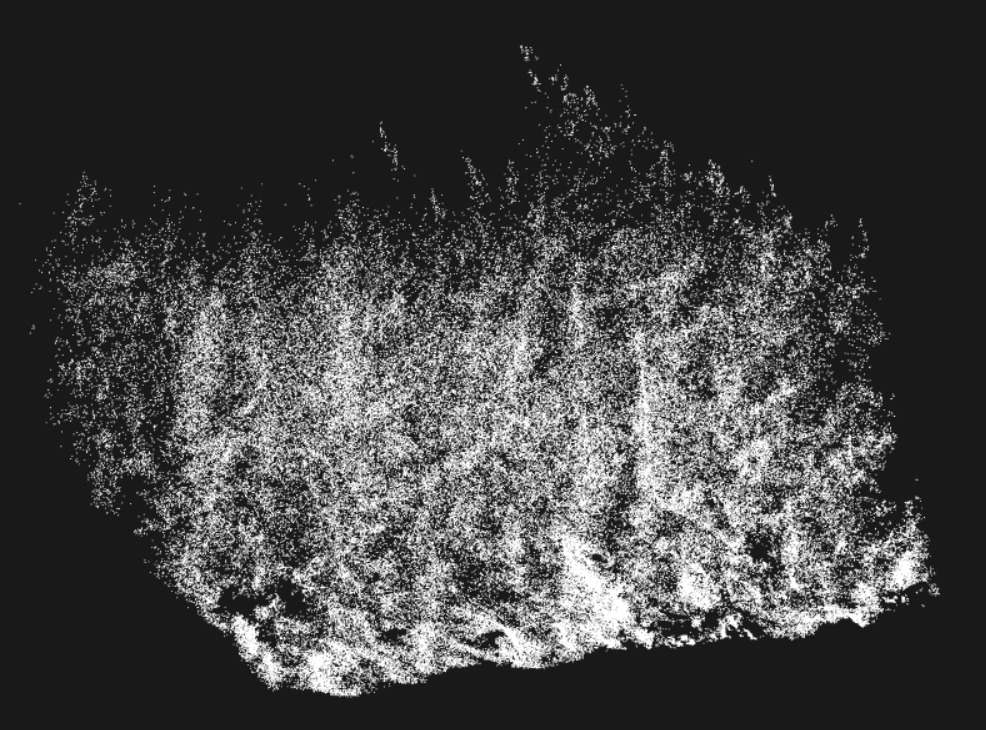}
  \caption{Point cloud visualisation of acquired data by VUX at SECIL1}
  
  \label{fig:Vux_SECIL1}
\end{figure}

\begin{figure} [htb]
\centering
    \includegraphics[width=0.25\textwidth]{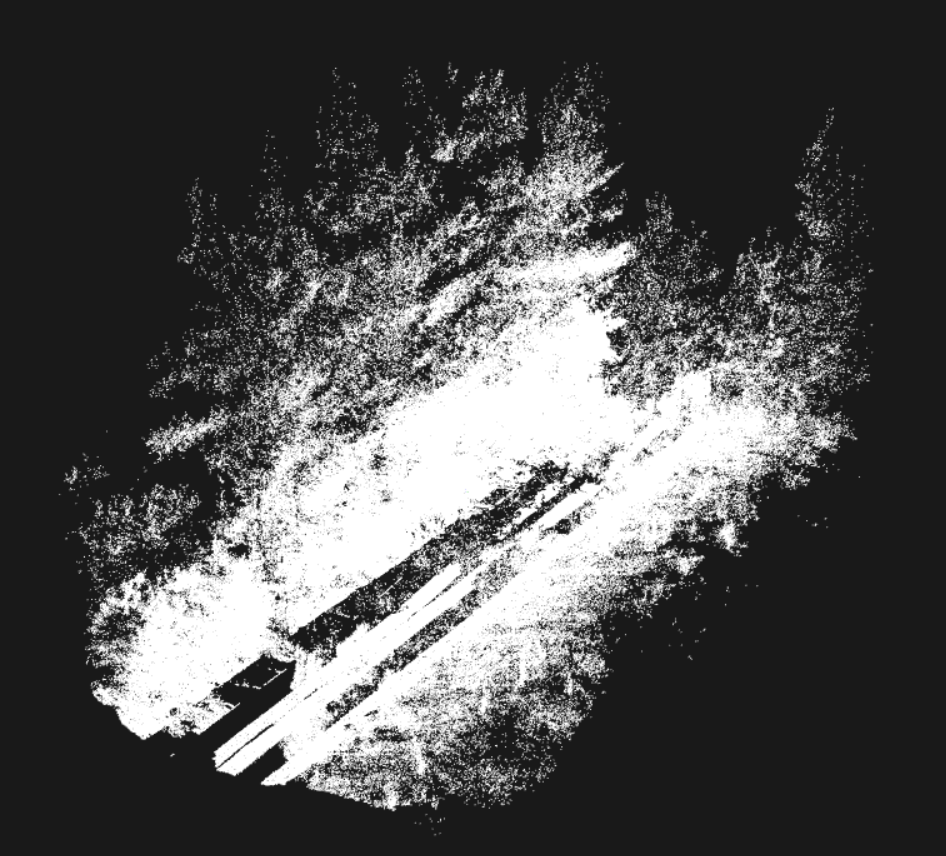}
  \caption{Point cloud visualisation of acquired data by VUX at SECIL2}
  \label{fig:Vux_SECIL2}
\end{figure}

This system was used to acquire point cloud data at the limestone quarry mentioned previously (figure \ref{fig:4.map}). From the area accessed by this system, only three locations were acquired by VUX and LCLS: SECIL1, SECIL2 and SECIL3. The figures \ref{fig:Vux_SECIL1} and \ref{fig:Vux_SECIL2} show the visualisation of the studied plot area for each location common with LCLS.

\subsubsection{Histograms of vegetation structure}

\begin{figure} [htb]
    \centering
    \includegraphics[width=0.4\textwidth]{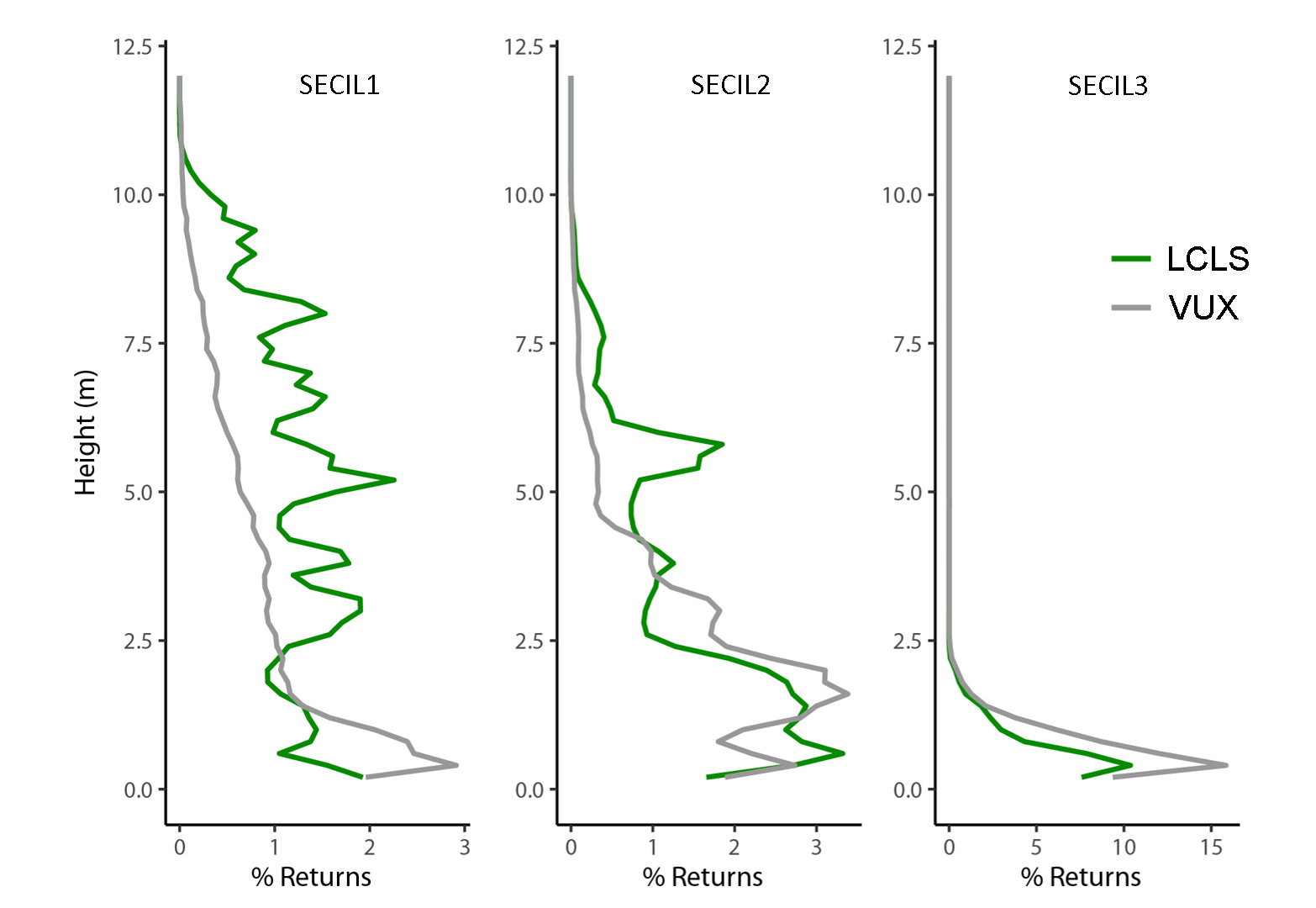}
    \caption{Histogram of vegetation structure of both VUX and VLP data for SECIL1, SECIL2 and SECIL3 locations.}
    \label{fig:4.histogram_vux}
\end{figure}

VUX point cloud data was also used to compute histograms of vegetation structure. These histograms used  the same method as the histograms computed for LCLS data. The figure \ref{fig:4.histogram_vux} presents the histograms data of both VUX and LCLS for each common location.

\subsection{Point density}

After clipping point cloud data from both VUX and LCLS data according to the field plots, point density for each the locations SECIL1, SECIL2 and SECIL3 were computed using the function grid\_density from lidR package. Using ArcMap v.10.6.1 (ESRI, 2019) point density was presented as number of points per grid cells of size 1m for each location.

\begin{figure} [htb]
    \centering
    \includegraphics[width=0.4\textwidth]{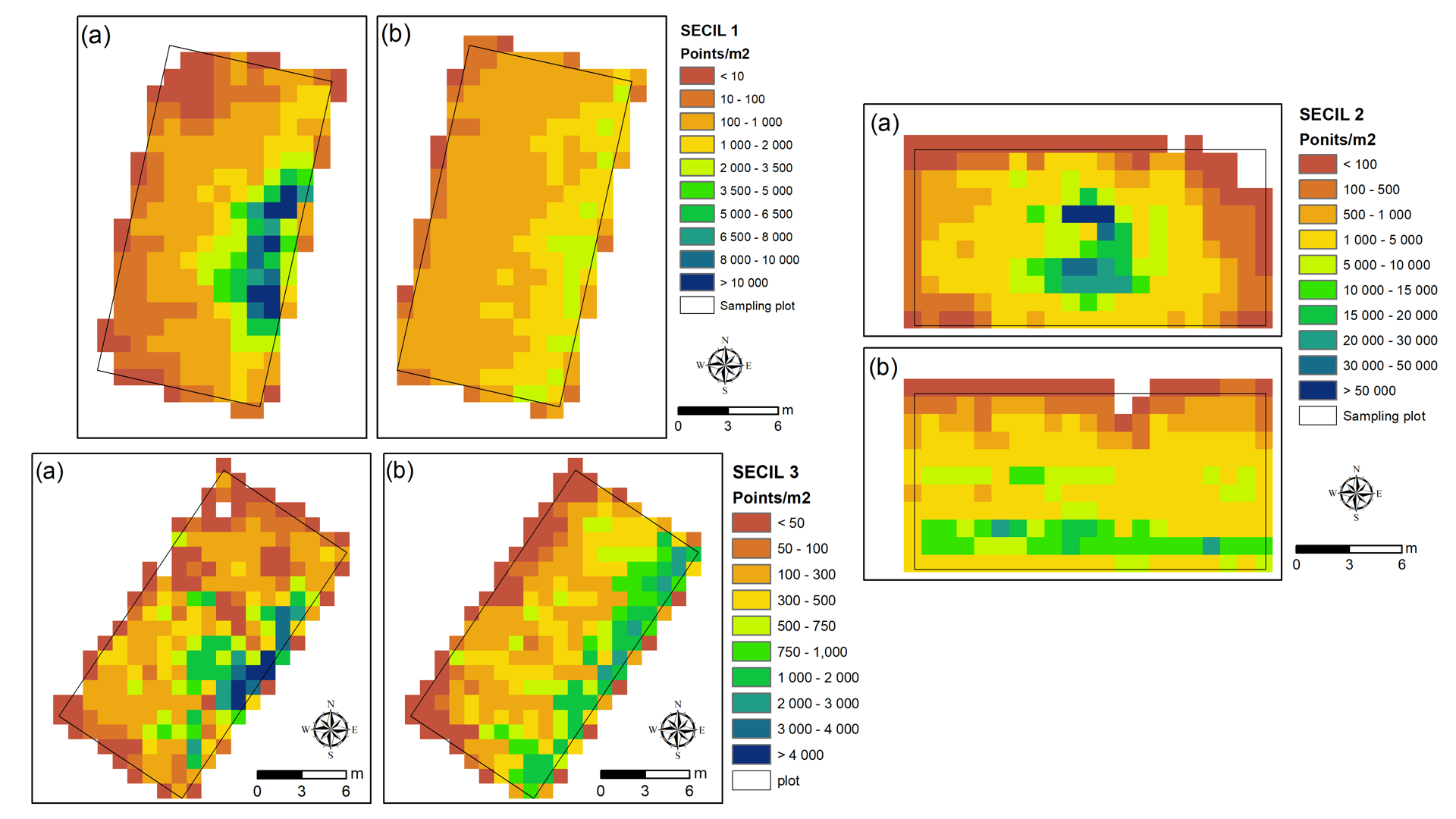}
    \caption{Point density representation for the locations SECIL1, SECIL2 and SECIL3. LCLS point cloud data corresponds to (a) and VUX point cloud data corresponds to (b).}
    \label{fig:point_density}
\end{figure}

Looking at the results in the figure \ref{fig:point_density}, it's clear that both system acquire more detailed information in areas closer to the sensor. From the results of LCLS, point density varies from the position of system in a circular pattern, with highest values being closest to the centre and decreasing with distance. For VUX, which is a mobile system, point density is the highest in regions closest to the path taken by the system and decreasing with distance to this path. LCLS shows a more advantageous instrument for acquisition of distances closer to the sensor, achieving a point density surpassing 50000 points/m${^2}$. 


\subsubsection{Point cloud data comparison with VUX data}

By observing the previous results, the developed LiDAR system was capable of detecting higher detail in vegetation structure with the tallest trees. The VUX system loses information in height, which shows the importance of the sensor location in mobile acquisitions. This system was not reaching the same 360{\textdegree} vertical plane acquisition as LCLS, which mostly likely was caused by occlusion effects when acquiring data from different positions of observation. For vegetation structures with lower height (such as as 2.5m shrubs height from SECIL3) both LCLS and VUX showed similar results.  Point density was higher for LCLS acquisition, with higher densities near the sensor. VUX had lower point density, which is expected since movement of the LiDAR sensor translates to decreased standing time for data acquisition \cite{beland2019promoting}.








\section{Conclusion and future work}



The main idea of the developed system was to change the orientation of a LiDAR sensor, that was capable of acquiring data with a horizontal field of view of 360\textdegree, in order to acquire distance measurements in a vertical plane. With complementary add-nos, such as a turn table and a step motor, sensors, such as an accelerometer, a gyroscope and a magnetometer, and a GPS, field of view of the LiDAR sensor was increased and allowed automation of point cloud correction and georeferencing. 


The system was tested in two types of environments, controlled and field tests. During performance analysis in the controlled environment, LCLS shown highly accurate distance measurements and point cloud resolution, acceptable accuracy in tilt calculations and some problems in determining angle difference with north, which may be caused due to calibration of the sensors used for comparison. After acquiring point cloud data during field tests, histograms of vegetation structure, represented as variation of percentage of total points from point cloud per ground height range, and the Shannon diversity index were computed which provided great insight in the environment vegetation complexity, complementing field measurements. When compared with a MLS, LCLS was considered the most appropriate tool for designated plot area. For acquisitions of larger areas the suggested system would be MLS.


In conclusion, LCLS showed promising results, having achieved the most important requirements. However, many improvements can be made, i.e. better sensor configuration and calibration methods, allow acquisition of multiple position within a plot area to form a point cloud with increased structure characterisation, and integration of RGB cameras with point cloud, which could used improve point classification (ground, non-ground, vegetation). For future work, it may also be interesting to develop a low cost mobile system, with similar requirements, in order to further increase the use of remote sensing technology in ecology studies.




%

\appendices


\section*{Acknowledgment}
The authors would like to thank SECIL-Group for allowing access to their limestone quarry in Setúbal, Portugal, and João Gomes Mota and Albatroz Engenharia SA. for sharing their point cloud data.



\ifCLASSOPTIONcaptionsoff
  \newpage
\fi



%

\bibliographystyle{IEEEtran}
\bibliography{references}

%






\end{document}